\documentclass[letterpaper, 10 pt, journal, twoside]{ieeetran}

\IEEEoverridecommandlockouts          

\usepackage{times}
\usepackage{graphicx}
\usepackage{amsmath}
\usepackage{mathabx}
\usepackage{algorithm}
\usepackage{algpseudocode}
\usepackage[T1]{fontenc}
\usepackage{array}
\usepackage{booktabs,ragged2e}
\usepackage{float}
\usepackage[flushleft]{threeparttable}
\usepackage{color,soul}
\usepackage{multicol}
\usepackage[bookmarks=true]{hyperref}
\usepackage{svg}
\usepackage{xcolor}
\usepackage{multirow}

\newcommand{\RNum}[1]{\uppercase\expandafter{\romannumeral #1\relax}}
\title{A Soft Continuum Robot with Self-Controllable Variable Curvature}

\author{Xinran Wang$^{1}$, Qiujie Lu$^{1,}$$^{2}$, Dongmyoung Lee$^{1}$, Zhongxue Gan$^{2}$, and Nicolas Rojas$^{1}$
\thanks{Manuscript received: September, 25, 2023; Revised December, 7, 2023; Accepted January, 1, 2024.}
\thanks{This paper was recommended for publication by Editor  Yong-Lae Park upon evaluation of the Associate Editor and Reviewers' comments. This work was supported in part by the Natural Science Foundation of China under Grant 52305013, and by the Shanghai Pujiang Program (22PJD006).}
\thanks{$^{1}$REDS Lab, Dyson School of Design Engineering, Imperial College London, 25 Exhibition Road, London, SW7 2DB, UK
{\tt\footnotesize  (xinran.wang20, q.lu17, d.lee20, and n.rojas)@imperial.ac.uk}}%
\thanks{$^{2}$Academy for Engineering and Technology, Fudan University, 200433, Shanghai, China.
{\tt\footnotesize   (qj\_lu,
ganzhongxue)@fudan.edu.cn}}

}

\begin{document}

\markboth{IEEE Robotics and Automation Letters. Preprint Version. Accepted January, 2024}
{Wang \MakeLowercase{\textit{et al.}}: A Soft Continuum Robot with Self-Controllable Variable Curvature} 




%

\maketitle

\begin{abstract}
This paper introduces a new type of soft continuum robot, called SCoReS, which is capable of self-controlling continuously its curvature at the segment level; in contrast to previous designs which either require external forces or machine elements, or whose variable curvature capabilities are discrete---depending on the number of locking mechanisms and segments. The ability to have a variable curvature, whose control is continuous and independent from external factors, makes a soft continuum robot more adaptive in constrained environments, similar to what is observed in nature in the elephant's trunk or ostrich's neck for instance which exhibit multiple curvatures. To this end, our soft continuum robot enables reconfigurable variable curvatures utilizing a variable stiffness growing spine based on micro-particle granular jamming for the first time. We detail the design of the proposed robot, presenting its modeling through beam theory and FEA simulation---which is validated through experiments. The robot's versatile bending profiles are then explored in experiments and an application to grasp fruits at different configurations is demonstrated. A narrated video detailing the work can be seen at \url{https://youtu.be/H6SCK0NjGpE}. 
\end{abstract}


\section{Introduction}
    \IEEEPARstart{S}{oft} continuum robots are in general flexible and adaptive. This feature allows them to be more easily actuated with less actuators than degrees of freedom and achieve more complex motions in constrained environments. However, the common approach for soft continuum robots to have more bending profiles to increase dexterity is adding more sections or segments~\cite{ santoso2021origami}\cite{mcmahan2005design}\cite{wang2022data}\cite{wooten2022environmental}. This strategy is effective but usually implies adding more actuators to control the continuum robot, while increasing design complexity.

Exploring stiffness change inside soft robots has become a popular approach to achieve a more dexterous workspace and higher external loads. The stiff-flop manipulator, for instance, explored granular jamming using coffee powder to allow stiffness change on silicone rubber-based soft continuum robots \cite{cianchetti2014soft}. Tendon-driven continuum robots can also utilize granular jamming to alter its curvature \cite{wockenfuss2022design}. Other alternative options for achieving uniform stiffness changes for each section of a soft continuum robot (i.e., the same stiffness across a section) include fiber jamming \cite{arleo2023variable}, scale jamming \cite{sadati2015stiffness}, shape memory alloy (SMA) \cite{jiang2020variable}, layer jamming \cite{clark2022malleable}, and pressurized fluidic compartments \cite{fras2023fluidic}.


\begin{figure}[htbp]
 \centering
 \includegraphics[width=0.95\linewidth]{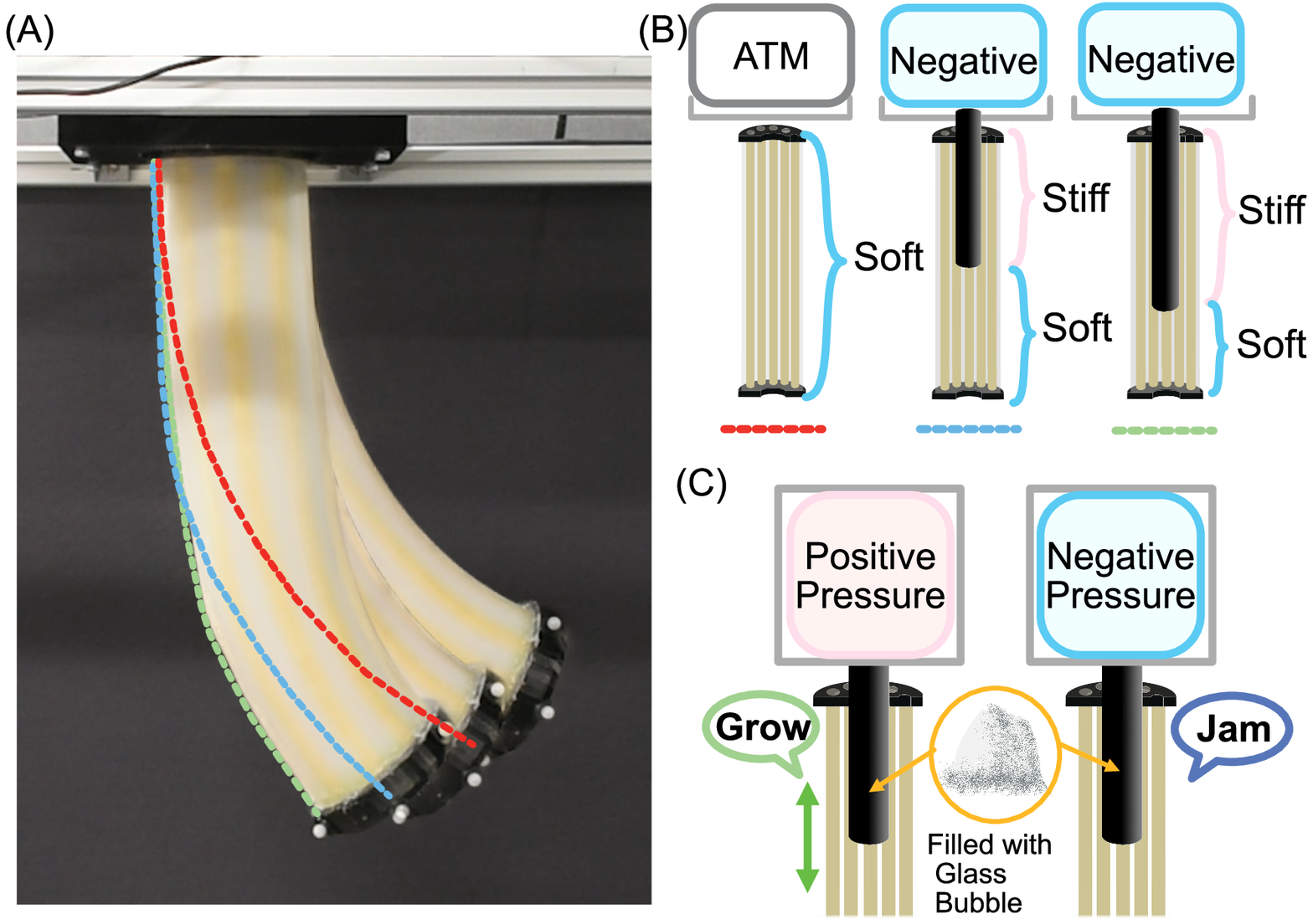} 
 \caption{(A) Illustration of controllable curvatures when the robot is pressurized at 250kPa. (B) Inner structure: The soft continuum robot has a hollow channel that allows the growing spine to travel inside. The jammed spine then creates a high stiffness inside while the remaining length is flexible. With the same control inputs, it creates different bending profiles. (C) Realizing concept: The growing spine filled with granules changes length using positive pressure and length control. Using negative pressure, it increases stiffness by granular jamming.}
 \label{fig:1}

\end{figure}

Recently, there has been a greater focus on a non-uniform stiffness change regardless of the number of actively actuated sections on the continuum robot, allowing an even more dexterous motion thanks to the resulting variable curvatures. Methods reported include using position locking with SMA, magnets, screws, or tendons along the robots \cite{pogue2022multiple}\cite{wang2022design}\cite{nakano2023robostrich}\cite{yang2020geometric}\cite{rao2023modeling}\cite{amanov2021tendon}. Other approaches use pre-programmed springs or disks and add them to the continuum robot to alter its local stiffness \cite{zhang2023preprogrammable} 
 \cite{ma2023inspired}. These methods for achieving variable curvatures have limitations due to the number of locking mechanisms or the need for manual configuration in advance. Soft robots with continuously changing stiffness are still an area of limited research, and current approaches often rely on rigid materials such as rods or steel tubes with patterns to alter stiffness \cite{kim2019continuously}\cite{zhao2020continuum}. 

\begin{figure*}[t!]
 \centering
 \includegraphics[width=0.9\textwidth]{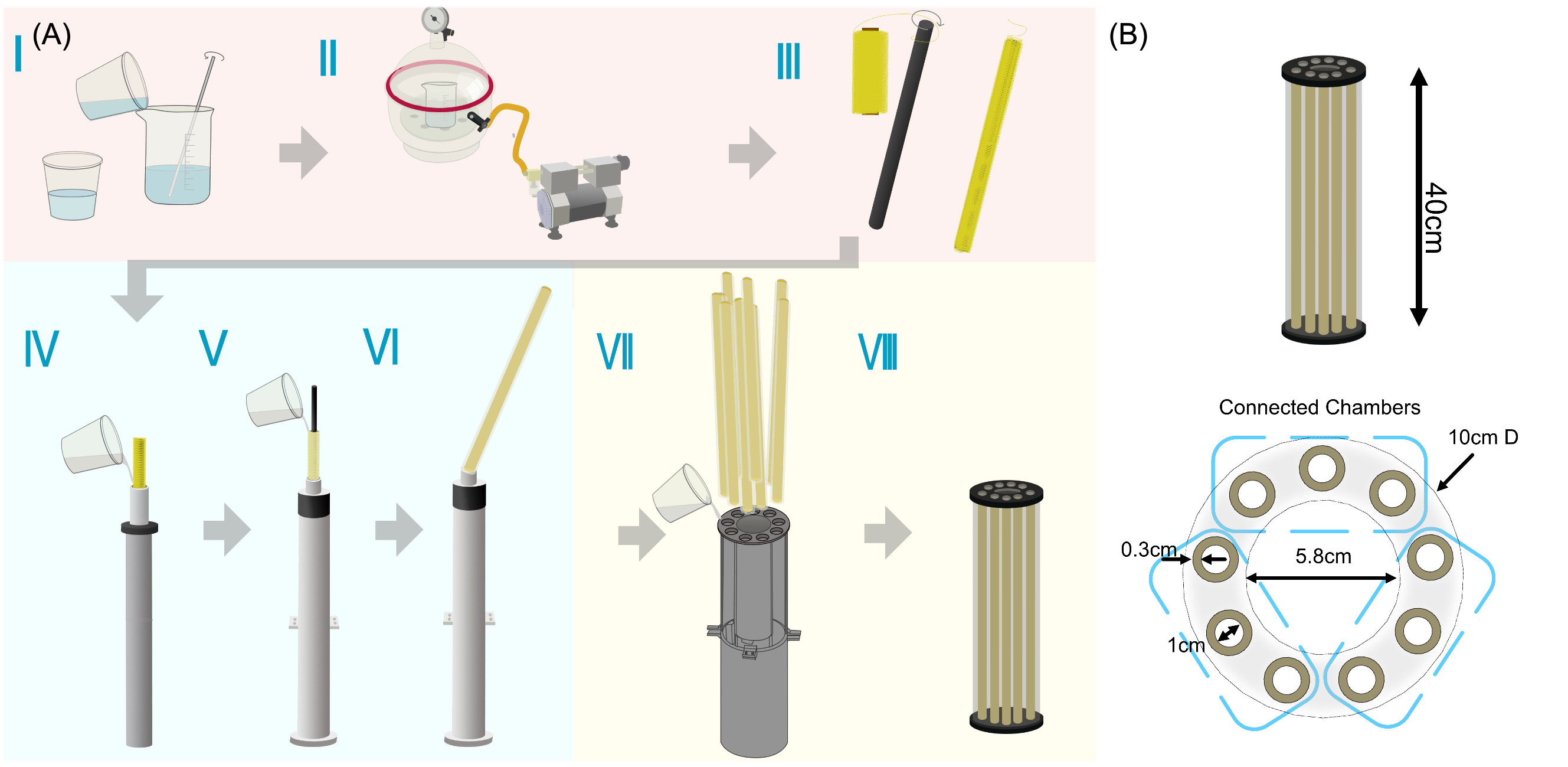} 
 \caption{(A) Fabrication of the soft continuum robot: silicone mixing, degassing, winding strings on the rod, molding silicone for each tube, assembling all the tubes for a single section continuum robot. (B) Dimension of the soft continuum robot. It has nine chambers, and each three of them are connected.}
 \label{fig:2}
\end{figure*}

Soft continuum robots that take inspiration from vine robots~\cite{hawkes2017soft}, known for their steering ability and growth capabilities, have recently been developed. These continuum robots feature an additional controllable parameter---length~\cite{greer2019soft}. The benefit of such soft continuum robot comes from the material everting from inside, allowing it to travel under different environments and locations while being robust to friction. Much progress has been made to adjust the stiffness change on such robots as well~\cite{jitosho2023passive}\cite{wang2020dexterous}. Utilizing layer jamming on the robot for instance, stiffness can vary along the body~\cite{do2020dynamically}. Despite these advantages, these lightweight growing continuum robots are designed for 2D ground navigation with limited 3D workspace. Our 3D soft continuum robot with variable stiffness and curvature capability stands out from other similar designs, as showcased in the advantages presented in Table \ref{table2}.





In this paper, we then propose a Soft Continuum Robot with Self-controllable variable curvature  (SCoReS) that leverages the growing characteristics of vine robots and micro-particle granular jamming effect, which is based on lightweight glass bubbles, to create a new type of soft continuum robot. Figure \ref{fig:1} presents the central concept to achieve variable bending curvature profiles; 
The main contributions of this research work as follows: 
\begin{itemize}
  \item We proposed a novel self-controllable variable curvature soft continuum robot design, which allows continuous stiffness change at the segment level.  
  \item Develop a growing spine that can grow and adjust its stiffness based on micro-particle granular jamming, along with a mechanism that allows the growing spine to control its length automatically. 
  \item The bending profiles of SCoReS have been simulated, and experimentally validated.
\end{itemize}


\begin{table}[t]
\centering
\caption{Comparison between existing soft continuum robots in 3D with variable stiffness or variable curvature capabilities}
\label{table2}
\begin{tabular}{c c c c c}
\hline
  & \multicolumn{4}{c}{Variable Stiffness Capability}  \\ \cline{2-5}
Research & Uniform & Localized & Continuous & Deployment \\ \hline
\cite{cianchetti2014soft}\cite{wockenfuss2022design}\cite{arleo2023variable}  & \multirow{2}{*}{Y} & \multirow{2}{*}{N} & \multirow{2}{*}{N} & \multirow{2}{*}{Various} \\
\cite{fras2023fluidic}\cite{wang2022design} & & & & \\
\cite{zhang2023preprogrammable}\cite{ma2023inspired} & Y & Y & N & Pre-configure\\ 
Our work & Y & Y & Y & Self-controlled\\ \hline
\end{tabular}
\end{table}

\section{Robot Design}

This section will discuss the comprehensive design of SCoReS robot. It has two main components: one variable stiffness growing spine and one soft continuum robot. The variable stiffness growing spine incorporates the ability to grow inside the hollow continuum robot body and varying degrees of stiffness automatically through granular jamming. By doing so, the single-section soft continuum robot can achieve the effect of multiple bending curvatures, allowing for greater flexibility and adaptability.

\begin{figure*}[t]
 \centering
 \includegraphics[width=0.9\textwidth]{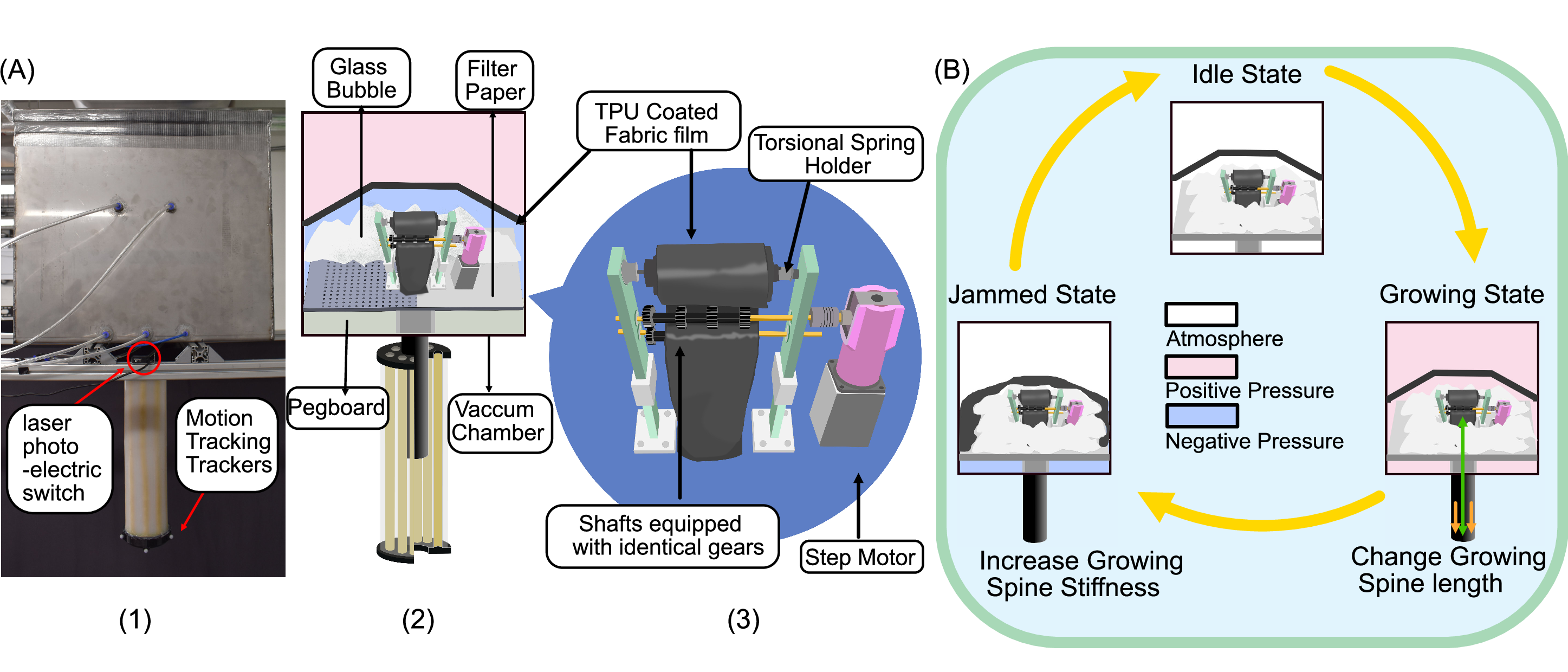} 
 \caption{System detailed design (A) 1. Experimental Setup, with a motion tracking system to track its end-effector positions and rotations. 2. Cross-section area to show the inner structure of the robot. Inside the steel box, airtight fabric divides the volume into two. The bottom volume contains glass bubbles and a length control mechanism. The filter paper on top of the pegboard prevents glass bubbles from going into the vacuum chamber when jamming. 3. Detailed length control mechanism: using a stepper motor for precise length control. (B) The state transitions to allow the SCoReS to reconfigure. }
 \label{fig:3}
\end{figure*}

\subsection{Continuum Robot Design}

The continuum robot comprises nine air chambers, with three of them interconnected for seamless three-dimensional movement. An electric proportional pressure valve regulates the pressure of each chamber. Furthermore, the robot features an inner channel that enables a growing robot to pass through.

We chose DragonSkin 10 (Smooth-on Inc.) as the material for our soft continuum robot, as it offers excellent stiffness levels comparable to EcoFlex 30 (Smooth-on Inc.) , and similar elongation properties. 

To prevent radical expansion of the air chamber, we have implemented separate constraints for each chamber. This way, the soft continuum robot can maintain a consistent inner diameter, regardless of its configuration, whether it's bent or straight. This feature ensures that the soft continuum robot has minimal inner radius change.

The process of creating the soft continuum robot is illustrated in Figure \ref{fig:2}(A). To begin, we mix two components of DragonSkin 10 silicone and put them in a vacuum container to degas. Then, we wind kevlar strings tightly around an aluminum rod in Step \RNum{3}. Next, we pour silicone onto the outside of the aluminum rod in Step \RNum{4}. Once the silicone is cured, we remove the aluminum rod, leaving the kevlar strings inside the molded silicone. In Steps \RNum{5} and \RNum{6}, we place a smaller aluminum rod in the center of the shell and pour silicone into the mold. After curing, we remove the small center aluminum rod and are left with the silicone air tube and perfectly embedded Kevlar strings. 

By integrating nine circular air channels into a new mold and merging them together with silicone, we can produce a soft continuum robot with an inner hollow channel. This process was previously only used for small surgical applications \cite{fras2015new}, but we have now successfully implemented it in a larger continuum robot measuring 40cm in length. Our design features an inner channel with an inner diameter of 5.8cm and an outer diameter of 10cm as the dimension of the continuum robot is in Figure \ref{fig:2}(B).

\subsection{Design of Variable Stiffness Growing Spine}

Creating multiple curvatures and providing variable stiffness in a single-section continuum robot has always been challenging. 

We are proposing a design of a self-controllable growing spine with variable stiffness inside an enclosed steel box. Our project first upgrades growing robots with a new length control unit for accurate position control.

Figure \ref{fig:3}(3) clearly illustrates the various components of the length control mechanism. These include a pair of shafts with gears, a pre-tensioned torsional spring holder for storing the growing robot material, and a NEMA23 stepper motor that comes with a planetary gearbox and an L-shaped connector. 

By applying positive pressure to the chamber, we create a constant exerting force on the air-tight ripstop TPU-coated fabric of the growing robot. As the stepper motor rotates, the fabric is released through the fixed-diameter geared shafts, enabling the growing spine to grow from the inside out without any high friction. By regulating the rotation of the stepper motor, we can effortlessly control the length of the growing robot (as seen in Video from 0:0:14s to 0:0:53s). 

The material is wrapped around a pre-tensioned torsional spring, which allows for retraction of the fabric while the stepper motor pulls it back. The step motor has been applied with air-tight glue and wrapped in TPU-coated fabric to prevent particle obstruction during operation. This mechanism ensures that the spine grows smoothly and consistently.

Glass bubble material has been utilized to enhance fluidity and prevent volume shrinking in the jamming process when applied with growing robots. Unlike conventional granular jamming techniques that use large and heavy particles, glass bubble consists of small and lightweight (0.2g/cc) round micro-size particles with a hollow spherical structure that promotes easy flow and movement while Capable of withstanding intense pressure during the jamming process \cite{bakarich2022pump}.

In Figure \ref{fig:3}(A-2),  the steel box is divided into two separate volumes by a flexible TPU-coated ripstop fabric. As shown in the pink section of Figure \ref{fig:3}(A-2), the pink volume is either pressurized or connected to the atmosphere. The blue-colored volume contains a length control mechanism, which is displayed in Figure \ref{fig:3}(A-3), along with glass bubble materials. To secure the length control mechanism in place, a pegboard is installed in the blue volume. Further, layers of industrial-grade filter paper and steel nets are placed on top of the pegboard to prevent the glass bubble from falling out. 

Finally, the green volume serves as a vacuum chamber, it connects with blue volume and growing spine, ensuring the safe and particle-free operation of the vacuum pump. When performing granular jamming, the growing spine stiffness will be significantly increased.

\begin{figure*}[t]
 \centering
 \includegraphics[width=0.9\textwidth]{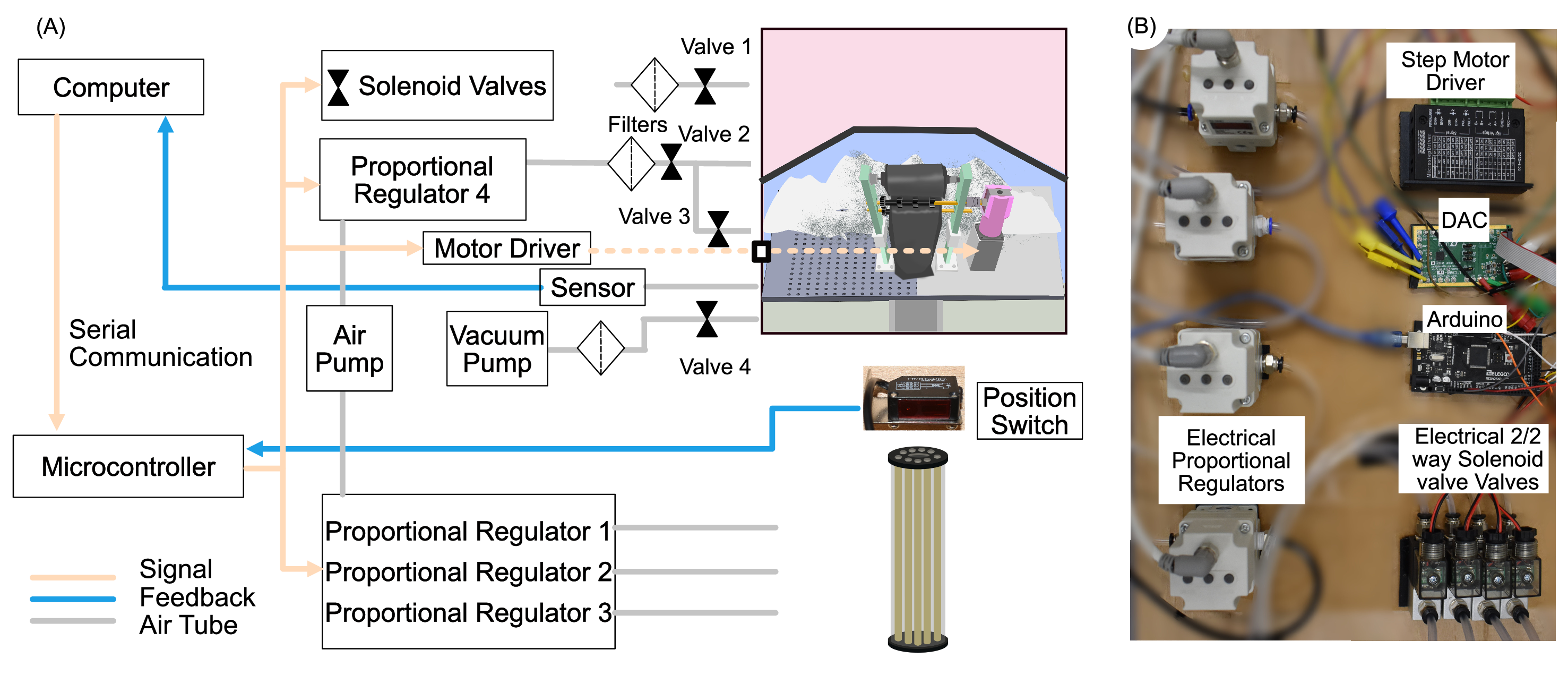} 
 \caption{(A) Control diagram of the overall system. (B) Detailed physical components: Electric Proportional Regulators, Step Motor Driver, Digital Analog Converter (DAC) for Proportional Regulators, Arduino and Solenoid Valves .}
 \label{fig:4}
\end{figure*}
\subsection{Overall System and Self-Controllable  workflow}
Figure \ref{fig:3}(A-1) displays the overall experimental setup. The soft continuum robot is positioned beneath the top-growing robot, and a laser photoelectric switch is placed between them for precise position calibration during the growing and retracting of the top-growing robot. The Opti-track motion tracking System is employed to track the end-effector constantly, providing real-time position and rotation information.

SCoReS undergoes three states in sequence: idle, growing, and jammed. In the idle state, the entire volume inside the cubic steel box is connected to the atmosphere. The robot neither grows nor retracts during this state.

The steel box in its cubic shape, is fully pressurized to 10kPa to facilitate the growth of the growing spine. As the spine grows, gravity helps the glass bubble to occupy all the empty spaces inside the growing spine. The internal pressure of the spine prevents any buckling during retraction. Furthermore, the fabric is pressurized to distribute glass bubbles evenly across the growing spine, ensuring they are not concentrated at the bottom. This allows more space for the glass bubble to move freely inside the growing spine.

When in the jammed state, the cubic steel box's top volume is connected to the atmosphere, while the rest of the volume is vacuumed through a vacuum pump. This effectively jams the growing robot in place and significantly increases its stiffness. When we control the soft continuum robot, it will bend in various curvatures due to alterations in its stiffness (as seen in the supplementary video from 0:0:54s to 0:01:25s). 

The SCoReS has the capability to adjust stiffness at various lengths and produce multiple bending curvatures by following the three states in order and utilizing different inner lengths during growth.

\subsection{Robot Control}

Figure \ref{fig:4}(A) illustrates that the control command is sent to the microcontroller via serial communication from a computer. To control of the bending of the soft continuum robot in three dimensions, we have employed three electric-proportional regulators. 

Four solenoid valves are in the system to control the two separate volume pressures inside the steel box. Valve 1 is linked to the atmosphere, while valve 2 is connected to an air pump via a proportional pressure regulator. Valve 3 is linked to the pink volume in the steel box, and valve 4 is linked to a vacuum pump.

When in the idle state, the steel box is connected to the atmosphere through the open of solenoid valve 1 while valve 2 remains closed. Valve 3 is open, valve 4 is closed.

During the growing state, the solenoid valve 1 and 4 are closed while valve 2 and 3 are open. The proportional regulator is adjusted to maintain a pressure of 10kPa to pressurize the steel box.

In the jammed state, the solenoid valve 1 is open to connect to the atmosphere while valve 2 and valve 3 are closed. Valve 4 is open to vacuum the steel box.

The computer receives pressure feedback from a pressure sensor measure range from -100kPa to 100kPa (SMC PSE543A-R06). The electric proportional pressure regulators we used are SMC ITV 2030. 

\section{Modelling and Simulation}

\subsection{Stiffness Modelling of Jammed growing Spine}
In order to accurately assess the stiffness of the jammed growing spine, we have employed the Euler-Bernoulli beam theory for mathematical modeling purposes. The beam model has been employed previously to check the variable stiffness capability in granular jamming \cite{jiang2014robotic}. Presently, we are evaluating this model on the jammed growing spine and integrating it with Finite Element Analysis for a comprehensive system evaluation.

The jammed growing spine has an airtight fabric outside and inside filled with material glass bubble. The jammed growing spine's stiffness can be represented as a cantilever beam problem for characterization purposes. In this scenario, one end of the beam is fixed while a point force acts upon the free end. The fundamental equation in beam theory serves as the basis for this analysis:
\begin{equation}
M = -E \times I \times \frac{d^2y}{dx^2}
\label{eq:1}
\end{equation}
Where: \( M \) is the bending moment. \( E \) is the Young's Modulus. \( I \) is the moment of inertia of the beam's cross-sectional area. \( y \) represents the deflection in the direction perpendicular to the beam length. \( x \) represents the position length from the fixed end along the beam.

The bending moment, in this case, is caused by a point load on the free end, for a cantilever length of $L$,  is:
\begin{equation}
M = -F \times (L - x)
\label{eq:2}
\end{equation}

Now combining equation \ref{eq:1} and \ref{eq:2}, we get relationship between point load $F$ and the deflection $y$:

\begin{equation}
\frac{d^2y}{dx^2} = \frac{F \times (L- x)}{E \times I}
\label{eq:3}
\end{equation}

Having two integrations over position $x$, and equations are constricted by two boundary conditions, on the fixed end, the \( \frac{dy}{dx} = 0 \) and when $x=0, y=0$, the equation can be simplified as:
\begin{equation}
y = \frac{F \times (3L-x) x^2}{6 \times E \times I}
\label{eq:4}
\end{equation}

The jammed growing spine has a circular shape, $I = \frac{\pi r^4}{4}$, rearranging equation \ref{eq:4} and substituting $I$, we can get $E$ when position is equal to the length of the beam $x = L$ :
\begin{equation}
E = \frac{4 \times F \times L^3}{3 \times \pi r^4 \times y }
\label{eq:5}
\end{equation}

\begin{figure}[t]
 \centering
 \includegraphics[width=8cm]{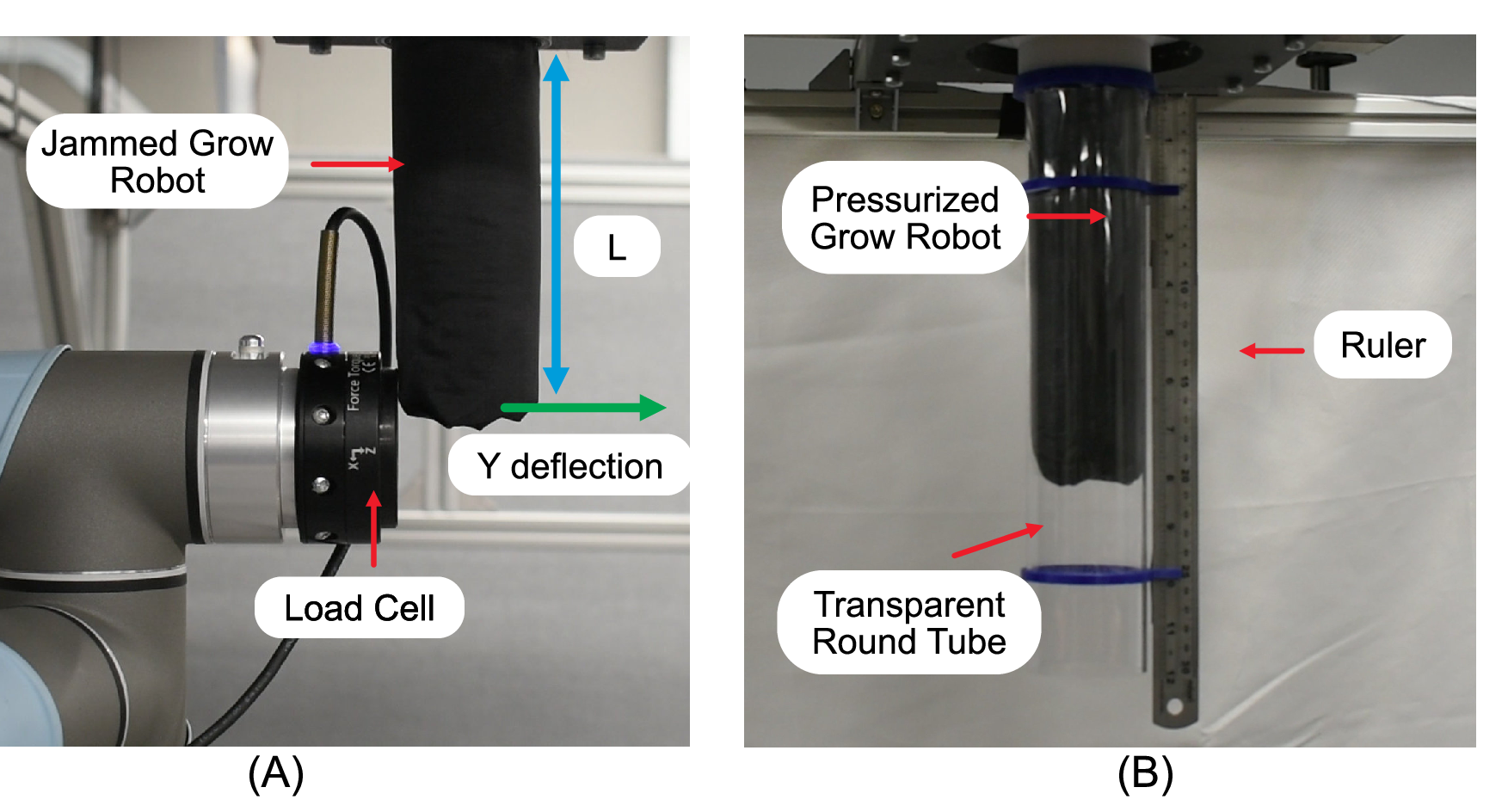} 
 \caption{(A) Experimental setup for bending experiments with load cell mounted on UR5. (B) Experimental setup for measuring length control with transparent tube and ruler.}
 \label{fig:5}
\end{figure}

\begin{figure}[t]
 \centering
 \includegraphics[width=0.9\linewidth]{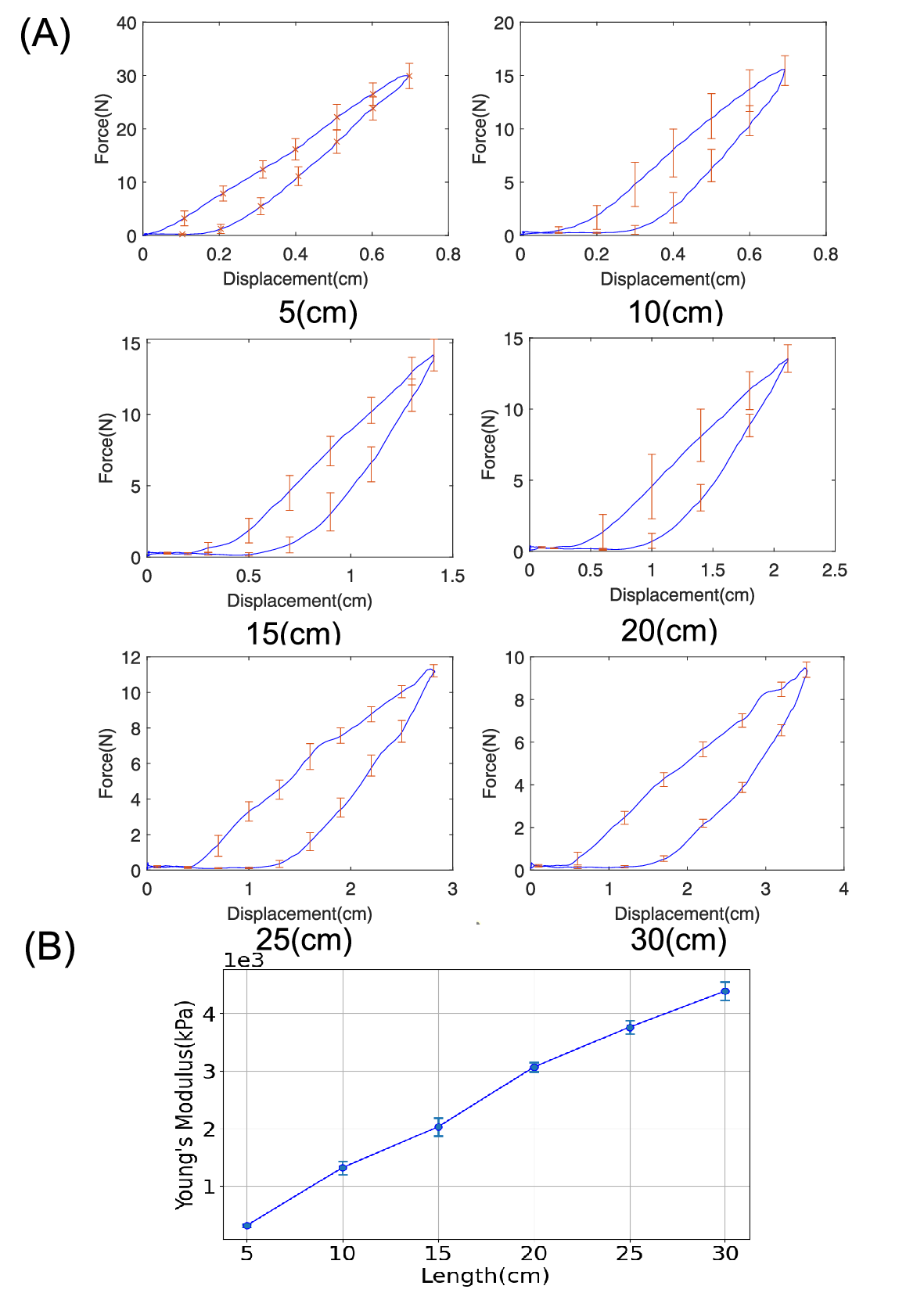} 
 \caption{(A) Bending Experiment during loading and unloading phrase for jammed growing robot at -70kPa and various lengths. (B) Calculated Young's modulus in kPa for the jammed growing spine at various lengths.}
 \label{fig:6}
\end{figure}

The overall experimental setup to measure the stiffness is provided in the \ref{fig:5}(A). For a jammed growing spine measured in length $L$, we have used a Universal Robots UR5 to push a distance of $Y$ in the direction perpendicular to the length direction of the growing robot. This model provides a fundamental basis for the following simulation analysis.

\begin{table}[t]
\centering
\caption{Length Growing Experiments}
\label{table1}
\begin{tabular}{c|c|c|c|c|c|c}
\hline
 Reference (cm) & 5 & 10 & 15& 20& 25 & 30 \\
\hline
Experiment (cm) & 5.25&10.53	&15.77	&21.06	&26.42	&31.43\\
\hline
Std Dev (cm)  & 0.30&	0.32&	0.31&	0.34	&0.39	&0.36 \\
\hline
\end{tabular}
\end{table}

The growing spine is able to grow at different lengths. To characterize the stiffness property of the jammed top-growing robot, we discretize the continuous length control problem into several quantitative segments in length for better comparison and understanding. We decided to evaluate the properties of SCoReS at these growing robot length segments: 5cm, 10cm, 15cm, 20cm, 25cm, and 30cm. 

\begin{figure*}[t!]
 \centering
 \includegraphics[width=0.9\textwidth]{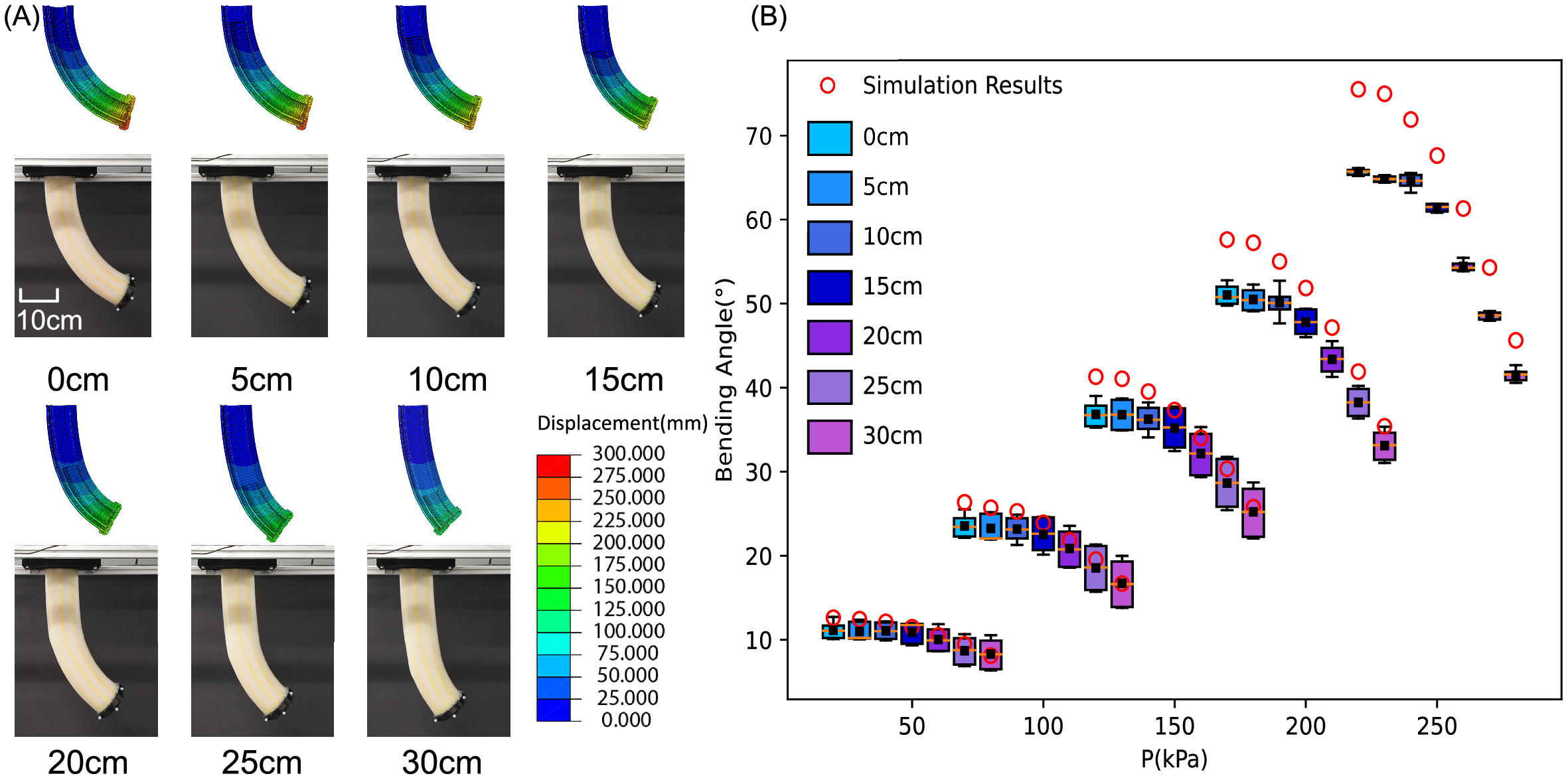} 
 \caption{FEA simulation and experimental results comparison for the SCoReS at different configurations. (A) The SCoReS overall bending profile comparison for different configurations at the pressure of 250kPa (B) Box plot for bending angle values comparison for the SCoReS pressurized at 50 kPa, 100 kPa, 150 kPa, 200 kPa, and 250 kPa with jammed growing spine segments ranging from 0 cm to 30 cm.}
 \label{fig:7}
\end{figure*}

For a jammed growing spine growing at different length $L$, UR5 push the jammed robot for a certain distance, and the load cell (ROBOTiq ft300-s) records the forces during the loading and unloading phase. The jammed spine is at -70kPa pressure. Figure \ref{fig:6}(A) shows the bending experiments results for the growing spine at different lengths. Based on the equation \ref{eq:5}, we are able to obtain the stiffness parameter for the jammed growing spine at different length segments.

The averaged stiffness values at different positions are  $0.318e^3$ kPa, $1.323e^3$ kPa, $2.032e^3$ kPa, $3.069e^3$ kPa, $3.763e^3$ kPa, $4.389e^3$ kPa for length segments of 5cm, 10cm, 15cm, 20cm, 25cm and 30cm. We can find as the length gets longer, the stiffness property also increases from Figure \ref{fig:6}(B). The main reason is that gravity also has an effect on the overall stiffness of the jammed-growing robot. The SCoReS is at a large scale. The length difference here is significant.  While the glass bubble material is light (0.2g/cc), the pressure at the bottom of the growing robot increases as the length gets longer. This, in return, increases the overall stiffness value.

We have also evaluated the length control accuracy of the growing spine through experiments. The setup in Figure \ref{fig:5}(B) was utilized for testing the length control accuracy at corresponding positions, and the reported results are listed in Table \ref{table1}. The average error was found to be $5.18\% $, while the maximum standard deviation error was 0.39 cm.

\begin{figure}[!ht]
 \centering
 \includegraphics[width=0.95\linewidth]{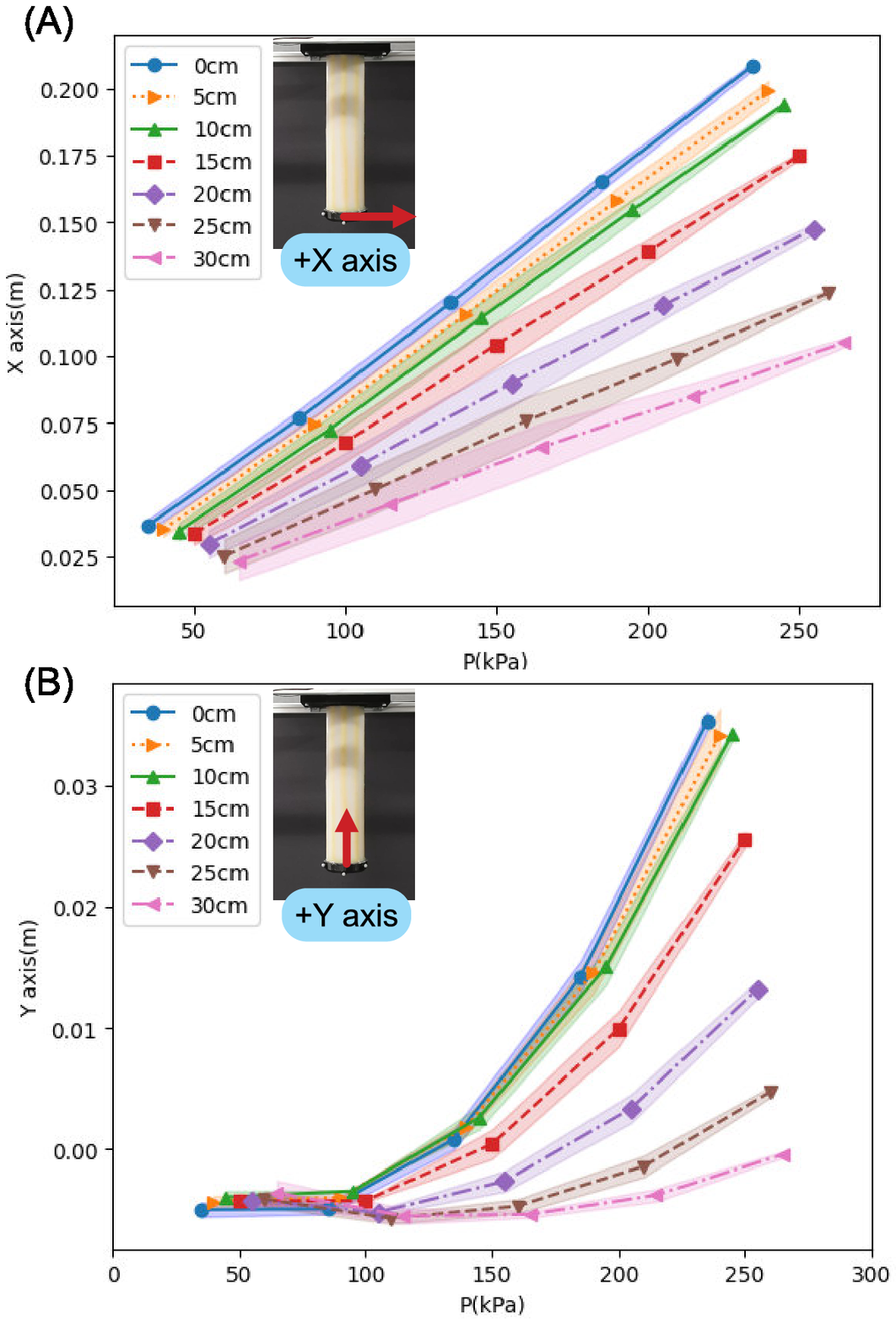} 
 \caption{Experiment results for bending length difference for SCoReS robot of different growing spine configurations (0cm, 5cm, 10cm, 15cm, 20cm, 25cm and 30cm growing spine length)} in A) X-axis direction, B) Y-axis direction.
 \label{fig:8}
\end{figure}

\subsection{Finite Element Analysis of the  Robot}

\begin{figure*}[t!]
 \centering
 \includegraphics[width=0.9\textwidth]{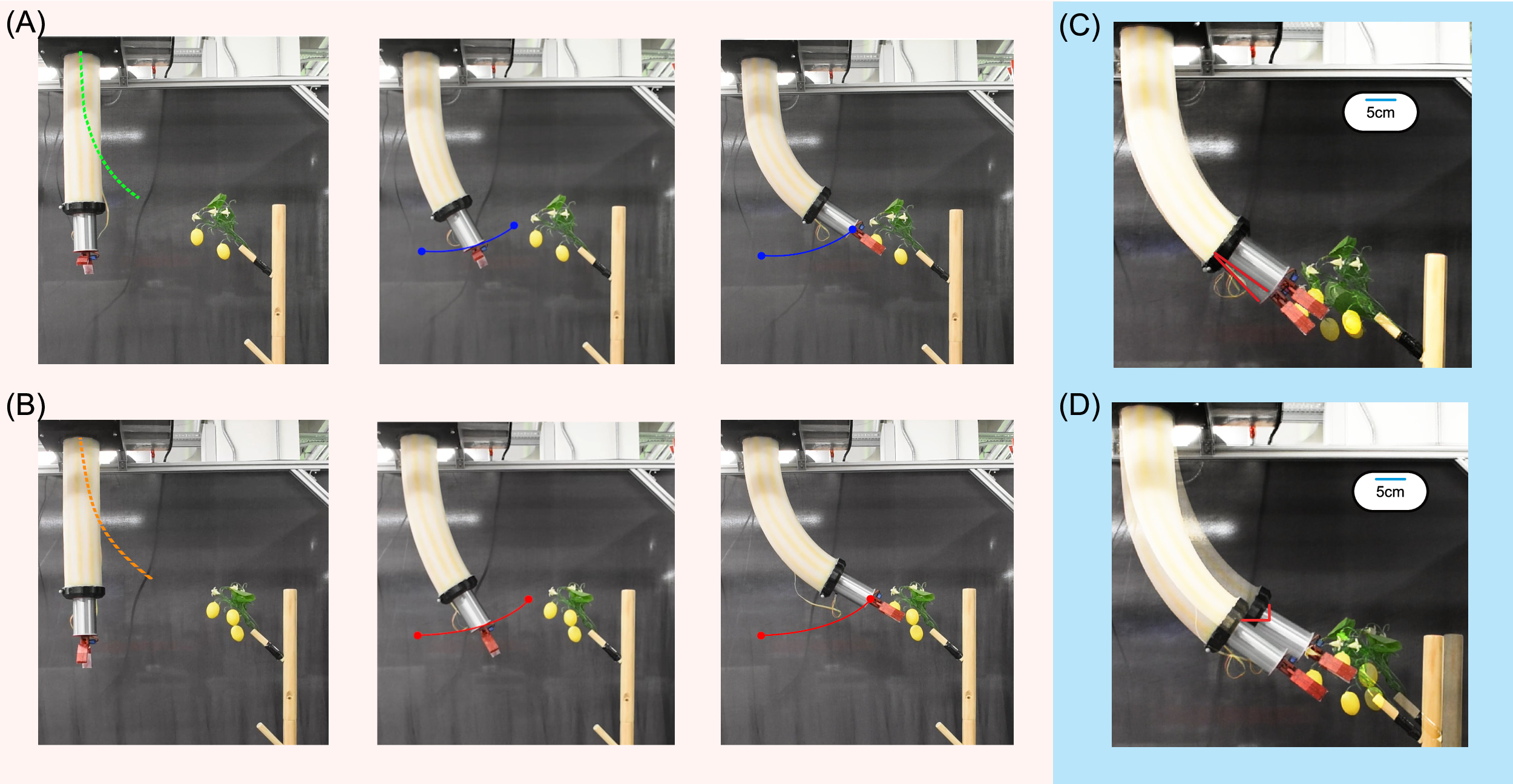} 
 \caption{Demonstrations of the SCoReS in various configurations in real-world applications.: 
A) 20cm jammed growing spine pressurized from 250kPa to 0kPa. 
B) 0cm jammed growing spine pressurized from 0kPa to 250kPa.  
C) Comparing a 20cm configuration pressurized at 250kPa to a 0cm configuration pressurized at 200kPa shows a difference in bending angle. 
D) Comparing a 20cm configuration pressurized at 250kPa to a 0cm configuration pressurized at 250kPa shows a difference in bending length.}
 \label{fig:9}
\end{figure*}
We incorporated the stiffness property of the jammed growing robot into our Finite Element Analysis (FEA) to model the interaction between the soft continuum robot's bending and the jammed growing robot.

In the previous section, we modeled the jammed top-growing robot as a cantilever beam. Using this modeling information, we create a cylindrical shape with the same diameter as the jammed-growing robot and place it inside the soft continuum robot. The chamber of the soft continuum robot is constrained similarly to the real-world case with kevlar strings inside the FEA simulation. The coefficient we used for DragonSkin 10 material is Neo-Hookean Model in Abaqus with $C_1$ = 42.5 kPa \cite{Polygerinos_Galloway_Wang_Connolly_Overvelde_Young}\cite{xavier2021finite}.

As shown in Figure \ref{fig:7}(A), the cross-section area of the SCoReS has a cylinder with different lengths inside the soft continuum robot in FEA. 
The image's color coding reveals the position displacement of the elements. At the 30cm configuration with a jammed growing spine inside, we observed a significant reduction in displacement and a shift in bending angle. With a portion of the soft continuum robot remaining stiff and the rest flexible, the inner jammed growing spine's length can be configured to alter the SCoReS's overall shape. This design makes the robot highly versatile, offering multiple different bending curvature profiles compared to the traditional single-section continuum robot.

\section{Experiment Evaluation}


\subsection{Bending Angle}

In order to validate our mathematical modeling and corresponding FEA simulation, we have conducted a comparison with our experimental results. As depicted in Figure \ref{fig:7}, our analysis focuses on the SCoReS with the inner jammed growing spine configuration at various lengths ranging from 0cm to 30cm. By comparing the bending angle from our experimental results with the FEA simulation, we have observed a close match between the two. Specifically, we found that the bending shape in FEA closely resembles the results of real experiments, particularly when the SCoReS is pressurized from 50kPa to 200kPa. It provides us with confidence in the accuracy of our mathematical modeling using beam theory and corresponding FEA simulation.

Upon closer inspection, we noticed that the maximum error occurred at 250kPa when the SCoReS's growing spine at 0cm. FEA predicted a bending angle of 75.48 degrees, while the experiment measured 65.64 degrees, with a maximum error of only 14.99\%. The error occurs without the interference of the jammed growing spines. Mentioning the material coefficients for DragonSKin 10 from the literature has a discrepancy with the real-world case.

In real experiments,  the SCoReS robot bent 41.50 degrees at 30cm and 250kPa pressure, compared to 65.64 degrees at 0cm configuration, a difference of 24.13 degrees. These findings provide us insight into the varying bending angles that the SCoReS robot can offer under different configurations.

\subsection{End Effector Position}

We have further examined the SCoReS robot bending profile with respect to the end-effector positions. Figure \ref{fig:8} illustrates the variations in the SCoReS robot extending length in both the X-axis and the Y-axis for different configurations. In particular, we observed that in the X-axis, there is a difference of 10.36cm in the distance between the robot with the inner growing spine at 0cm segment (20.87cm) and the robot at 30cm segment (10.51cm) when at 250kPa. 

In the Y-axis direction, there is a variation of 3.57 cm at 250kPa between the robot at 0cm configuration and the robot at 30cm configuration. Although this is less than the changes observed in the X-axis, the robot at the configuration of 30cm exhibits a much more stable motion in the Y-axis direction being pressurized from 0kPa to 250kPa.

\subsection{Real World Application Demo}

Furthermore, we showcase a demo of a real-world application by mounting a gripper on the SCoReS robot. As illustrated in Figure \ref{fig:9}, a clear comparison is made between the robot with the inner growing spine at 0cm segment (Figure \ref{fig:9}(A)) and the robot at the configuration of 20cm (Figure \ref{fig:9}(B)). It is evident from Figure \ref{fig:9}(A) that half of the continuum robot remains rigid in contrast to Figure \ref{fig:9}(B) even when both are pressurized to 250kPa. Moreover, the end-effector path highlights the distinct bending profiles. 

In Figure \ref{fig:9}(C), we compare the SCoReS robot with 0 cm configuration pressurized to 200kPa and the robot with 20cm pressurized to 250kPa. The robot's end effectors remain in a similar position, but the robot displays a more versatile bending profile with a 12-degree divergence in bending angle. This proves the effectiveness of the robot design.
Last but not least, at a pressure of 250kPa, Figure \ref{fig:9}(D) shows varying extension lengths for the two mentioned configurations, making it valuable in restricted spaces.

\section{Conclusion} 

In this study, we proposed the design of SCoReS that can be self-controllable to achieve a versatile bending profile. 
The jammed growing spine design has both characteristics of growing and jamming, making it unique in the application of variable curvature soft continuum robots. We used beam theory to model the jammed growing spine and conducted FEA analysis on the SCoReS. Our approach proved to be appropriate, as demonstrated by the close match between FEA and real experiments. Our experiments also showed significant changes in the bending profile, including bending angles and end-effector positions. In addition, The SCoReS's real-world demonstrations illustrate its ability to execute complex motions, such as varying the bending angle at the same end-effector location and shortening its bending length within a constrained environment. In the future, we plan to test the capability of the robot when jammed in different bending configurations instead of straight, and improve the design by adding fabric sleeves to anchor transportation units on the SCoReS. We'll also explore control strategies based on the calibrated silicone elastic models to fully utilize its bending profiles in constrained environments.

\bibliography{references}
\bibliographystyle{ieeetr}

\end{document}